\documentclass[sigconf]{acmart}

\usepackage{colortbl}
\usepackage{makecell}
\newcommand{\ecgstart}{\textsc{ECG-Start}}
\newcommand{\ecgtok}[1]{\textsc{ECG}_{#1}}

\setcopyright{acmlicensed}
\copyrightyear{2026}
\acmYear{2026}
\acmDOI{XXXXXXX.XXXXXXX}
\acmConference[RespMultimodal @ KDD '26]{RespMultimodal 2026: First International Workshop on Responsible Multimodal Foundation Models for Knowledge Discovery, held in conjunction with ACM SIGKDD 2026}{August 9--10, 2026}{Jeju, Korea}
\acmISBN{978-x-xxxx-xxxx-x/26/MM}

\begin{document}

\title{UniECG: Understanding and Generating ECG in One Unified Model}

\author{Jiarui Jin}
\authornote{Equal contribution.}
\affiliation{%
  \institution{Peking University}
  \city{Beijing}
  \country{China}}

\author{Xiuhan Zhang}
\authornotemark[1]
\affiliation{%
  \institution{Shanghai Ocean University}
  \city{Shanghai}
  \country{China}}

\author{Haoyu Wang}
\authornotemark[1]
\affiliation{%
  \institution{Peking University}
  \city{Beijing}
  \country{China}}

\author{Xingliang Wu}
\affiliation{%
  \institution{the Second Hospital of Tianjin Medical University}
  \city{Tianjin}
  \country{China}}

\author{Xiang Lan}
\affiliation{%
  \institution{National University of Singapore}
  \city{Singapore}
  \country{Singapore}}

\author{Jun Li}
\affiliation{%
  \institution{Peking University}
  \city{Beijing}
  \country{China}}

\author{Hongyan Li}
\authornote{Corresponding authors.}
\affiliation{%
  \institution{Peking University}
  \city{Beijing}
  \country{China}}
\email{leehy@pku.edu.cn}

\author{Shenda Hong}
\authornotemark[2]
\affiliation{%
  \institution{Peking University}
  \city{Beijing}
  \country{China}}
\email{hongshenda@pku.edu.cn}

\renewcommand{\shortauthors}{Jin et al.}

\begin{abstract}
Electrocardiogram (ECG) interpretation is a fundamental skill in medical education, yet students often need more than static examples to connect waveform evidence with diagnostic reasoning. This paper presents UniECG as a step toward interactive ECG education. UniECG supports two complementary learning interactions: given an ECG signal or image, it generates an evidence-based explanation; given a textual learning objective, it generates a corresponding ECG signal example for case-based learning. The model follows a two-stage design. First, it learns grounded ECG explanation from ECG signal--image--text data. Second, it introduces special ECG generation tokens and aligns their hidden representations with a pretrained text-conditioned ECG diffusion model, enabling controllable signal-level ECG generation. We evaluate UniECG through grounded ECG explanation and generation-oriented qualitative analysis, examining its potential to support explanation and case-based learning. UniECG is intended as an educational aid and a research step toward interactive AI-assisted ECG learning, rather than a clinically validated diagnostic system.
\end{abstract}



\maketitle

\section{Introduction}

\begin{figure}[t]
  \centering
  \includegraphics[width=\linewidth]{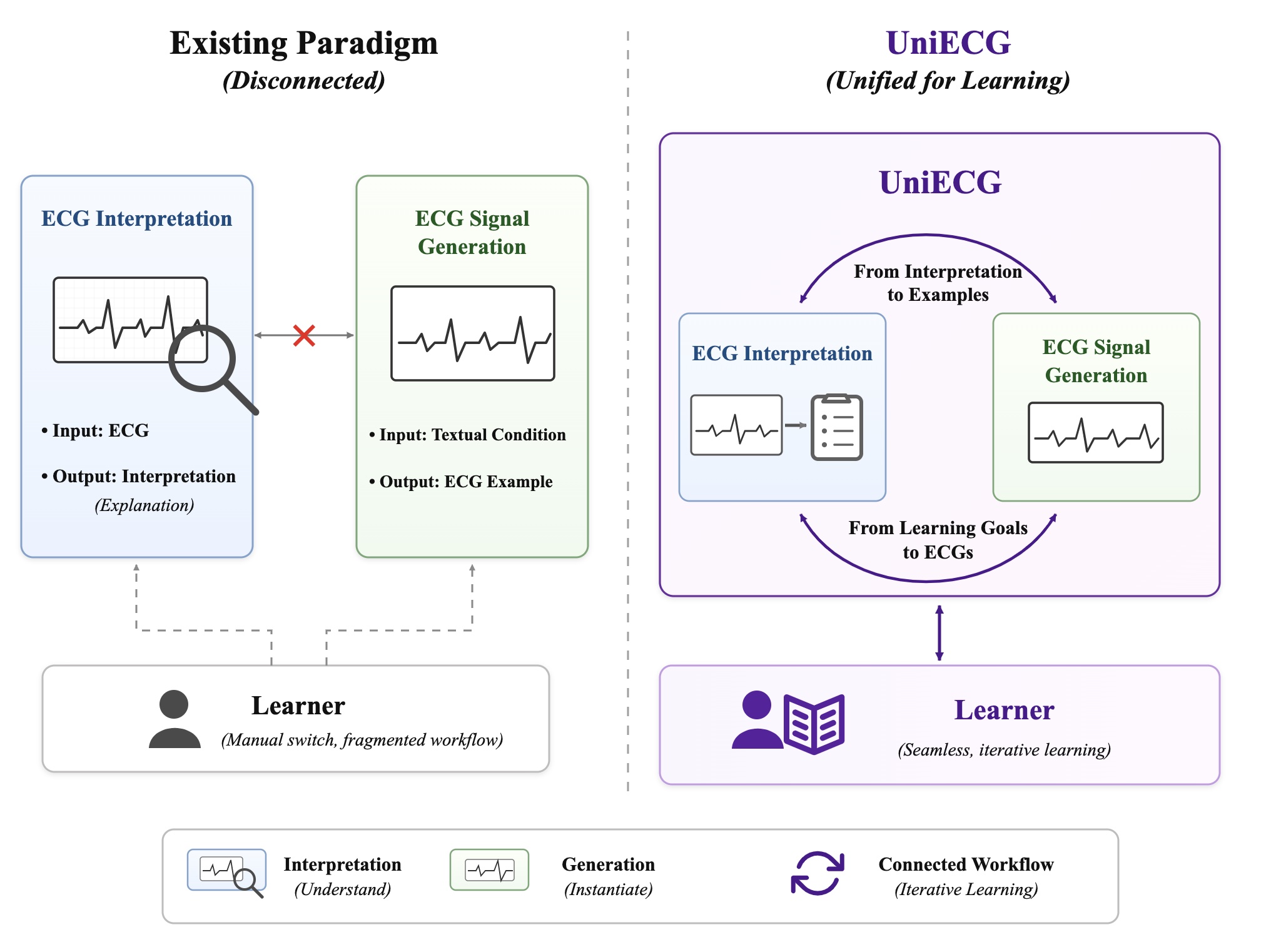}
  \caption{Motivation and overview of UniECG. Existing AI-assisted ECG learning separates ECG interpretation from ECG signal generation, creating fragmented learning workflows. UniECG unifies both modes into a closed-loop system for iterative ECG learning.}
  \label{fig:intro}
\end{figure}

Electrocardiogram (ECG) interpretation is a core skill in medical education and clinical training. However, ECG interpretation remains a formidable challenge for medical students. A systematic review and meta-analysis indicated that the baseline accuracy for ECG interpretation among students was a mere 42.0\% prior to supplemental training~\cite{10.1001/jamainternmed.2020.3989}. Furthermore, clinical competence statements from the ACC/AHA suggest that most practitioners require the interpretation of at least 500 ECGs under expert supervision to attain clinical proficiency~\cite{doi:10.1161/circ.104.25.3169}. Accordingly, how to integrate AI technologies to support ECG learning has become a research question of both practical importance and educational significance. Learning to read ECGs requires more than recognizing diagnostic labels: students must understand how rhythm, intervals, waveform morphology, and lead-specific findings are described and organized into a coherent interpretation. In practice, learners often encounter a tracing, read or request an interpretation, and then need additional examples that illustrate the same clinical concept. This makes ECG education naturally case-based and iterative rather than a one-shot diagnostic task.

\begin{figure*}[t]
  \centering
  \includegraphics[width=\textwidth]{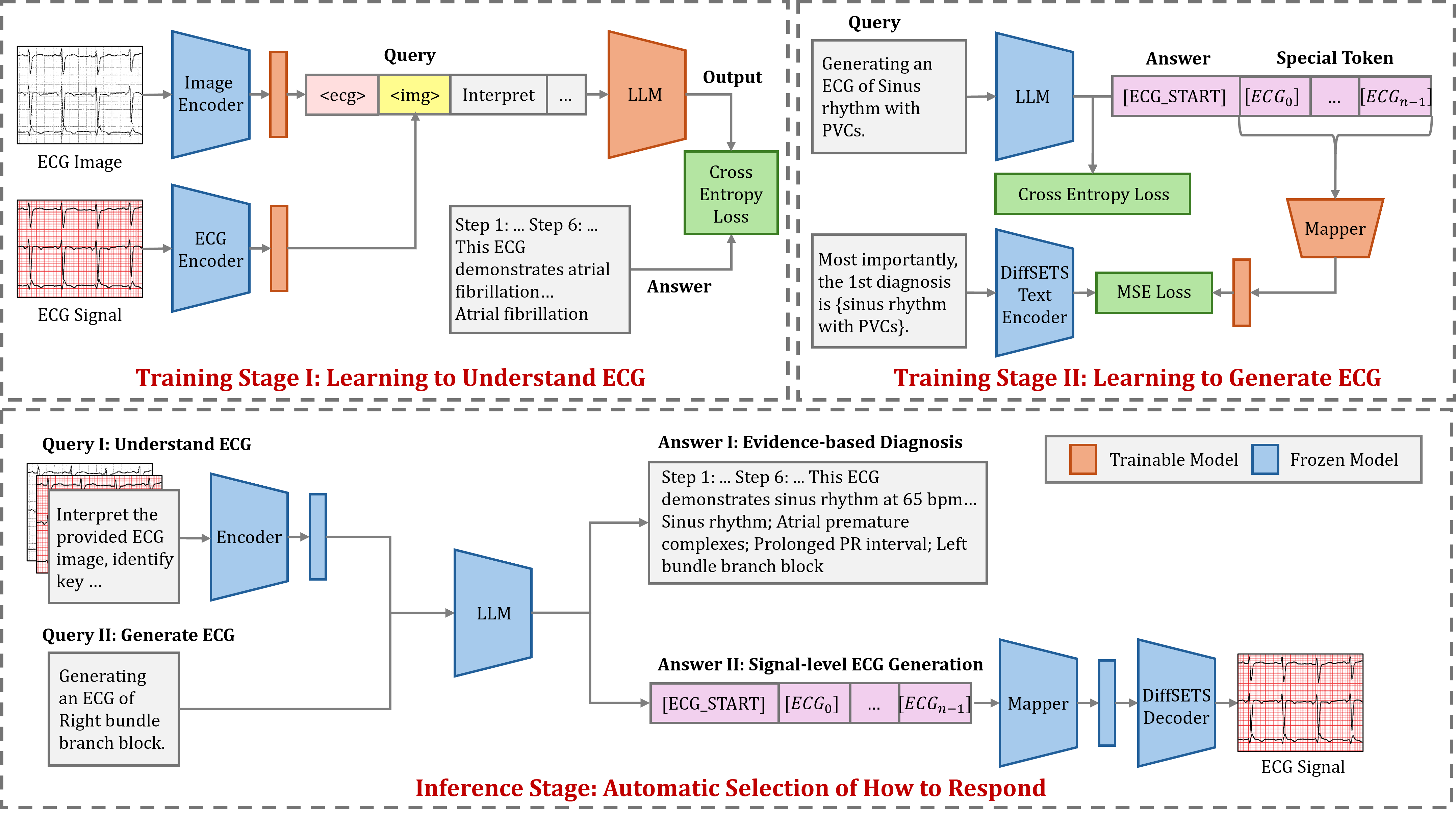}
  \caption{Framework of UniECG.}
  \label{fig:framework}
\end{figure*}

This setting creates two connected requirements for AI-assisted ECG education. First, a system should provide detailed, evidence-oriented interpretations that make the reasoning process explicit. Second, it should help construct condition-specific ECG signal examples that instantiate the same learning objective under controllable textual conditions. The two requirements are coupled: interpretation identifies the evidence and concepts that should be learned from an ECG, while ECG signal example construction provides additional examples for reinforcing or illustrating those concepts.

Existing AI systems do not fully support this workflow. Deep learning has substantially advanced automated ECG analysis~\cite{Opportunitieschallenges_Hong_2020,Deeplearning_Liu_2021}, but most prior work is optimized for diagnostic tasks rather than education. General unified multimodal large language models (MLLMs), such as GPT and Gemini, provide flexible understanding-and-generation interfaces but remain unreliable for ECG-centered learning: they may hallucinate diagnostic explanations, and their generated ECG signal examples may contain waveform patterns that are inconsistent with the requested clinical condition. Recent ECG MLLMs improve ECG interpretation~\cite{TeachMultimodal_Liu_2024,ECG-ChatLarge_Zhao_2025,GEMEmpowering_Lan_2025, ECG-R1Protocol-Guided_Jin_2026}, and ECG generative models synthesize ECG signals from textual or clinical conditions~\cite{SyntheticECG_Zanchi_2025,DiffuSETS12-Lead_Lai_2025}; however, these capabilities are usually studied separately, leaving explanation and ECG signal example construction disconnected.

Motivated by this gap, we present UniECG as a step toward interactive ECG education. UniECG connects two complementary learning modes in a single interface: given an ECG signal or image, it produces an evidence-based interpretation; given a textual learning objective, it generates a corresponding ECG signal example. The goal is not only to cover understanding and generation as separate capabilities, but to expose them as connected operations in an educational workflow.

Technically, UniECG follows a two-stage training strategy. The first stage learns evidence-based ECG explanation from ECG signal--image--text data, enabling the model to ground textual responses in ECG evidence. The second stage introduces special ECG generation tokens and aligns their hidden representations with the input space of a pretrained text-conditioned ECG diffusion model, enabling controllable ECG signal example generation. During inference, UniECG can respond with either an explanatory answer or generation tokens, depending on the user's educational request.

Our contributions are summarized as follows:
\begin{itemize}
  \item We introduce UniECG, a unified ECG model designed for interactive ECG education, which concurrently supports evidence based ECG explanation and text-conditioned ECG signal example generation.
  \item We adapt a two-stage training strategy for the educational workflow. UniECG first learns evidence-based ECG interpretation from signal--image--text data, and then injects text-to-ECG generation ability through latent-space alignment with a pretrained ECG diffusion model.
  \item We study UniECG through grounded ECG explanation evaluation and generation-oriented qualitative analysis, showing its potential to connect explanatory learning and case-based learning in a single ECG education workflow.
\end{itemize}

\section{Related Work}

\subsection{ECG Interpretation}

Early deep learning studies on ECG mainly focused on automated diagnostic tasks, including disease classification, report generation, and representation learning from ECG signals~\cite{Opportunitieschallenges_Hong_2020,Deeplearning_Liu_2021}. More recent ECG MLLMs extend this line of work by connecting ECG signals or images with natural-language outputs. For example, PULSE studies ECG image comprehension with instruction tuning~\cite{TeachMultimodal_Liu_2024}, ECG-Chat explores ECG-language modeling for cardiac disease diagnosis~\cite{ECG-ChatLarge_Zhao_2025}, and GEM introduces grounded ECG understanding with both time-series and image modalities~\cite{GEMEmpowering_Lan_2025}. ECG-R1 further emphasizes reliable interpretation through a protocol-guided and modality-agnostic MLLM design~\cite{ECG-R1Protocol-Guided_Jin_2026}. Together, these works move ECG analysis beyond label prediction toward natural-language interpretation, evidence grounding, and hallucination reduction. Our work builds on this direction, but focuses on an educational setting where detailed ECG interpretation is paired with ECG signal example generation for case-based learning.

\subsection{ECG Generation}

ECG generation has been studied for data synthesis, augmentation, and controlled waveform construction. Earlier methods include GAN-based and simulator-based synthesis, while recent work increasingly adopts diffusion models for flexible signal generation~\cite{SyntheticECG_Zanchi_2025,Diffusion-basedConditional_Alcaraz_2023}. Text-conditioned generation further connects clinical descriptions with ECG synthesis: Text-to-ECG studies 12-lead ECG synthesis from clinical reports~\cite{Text-to-ECG12-Lead_Chung_2023}, DiffuSETS generates 12-lead ECGs conditioned on clinical text reports and patient-specific information~\cite{DiffuSETS12-Lead_Lai_2025}, and SE-Diff incorporates physiological simulator constraints and experience-based retrieval to improve fidelity and semantic alignment~\cite{SimulatorExperience_Wang_2025}. These methods provide useful tools for constructing synthetic ECG signal examples, but they are typically independent from ECG interpretation models. In an educational setting, this separation limits how generated ECG signal examples can be used: a generated ECG may illustrate a condition, but it is not accompanied by a detailed interpretation that explains the relevant waveform evidence; conversely, an interpretation model may describe evidence in an observed ECG, but it cannot construct additional ECG signal examples that instantiate the same learning objective. UniECG addresses this gap by connecting evidence-based ECG interpretation with text-conditioned ECG signal example generation in a single workflow.

\section{Method}

UniECG is designed to support two educational interactions within a single model interface: explaining an observed ECG and generating an ECG signal example from a textual learning objective. The model is trained in two stages. The first stage learns grounded ECG explanation from ECG signal--image--text data, while the second stage equips the model with controllable ECG generation by aligning special ECG tokens with a pretrained ECG diffusion model.

\subsection{Learning Evidence-Based ECG Interpretation}

Given an ECG signal $s$, an ECG image $x$, and a textual response $y$, the first stage aims to fine-tune a large language model so that it can generate clinically meaningful explanations grounded in ECG evidence. We use ECG-Grounding~\cite{GEMEmpowering_Lan_2025}, which contains evidence-based ECG diagnostic dialogues derived from multimodal ECG data. The textual response is tokenized as $(y_1, \ldots, y_T)$. Following the decoupled modality-injection design of ECG-R1~\cite{ECG-R1Protocol-Guided_Jin_2026}, the interpretation stage uses separate signal and image pathways. A time-series encoder extracts ECG signal representations, and a visual encoder extracts ECG image representations. Two independent lightweight projectors then map these modality-specific representations into the LLM embedding space:
\begin{align}
e^{s} &= E_{s}(s), & z^{s} &= P_{s}(e^{s}), \\
e^{x} &= E_{x}(x), & z^{x} &= P_{x}(e^{x}),
\end{align}
where $z^{s}$ and $z^{x}$ are token blocks in the LLM embedding space. The projected ECG signal token block $z^{s}$ is inserted at an explicit ECG placeholder, while the image token block $z^{x}$ is inserted at the image placeholder. This decoupled injection allows the interpretation model to be trained with signal-only, image-only, or signal--image inputs, depending on which modality placeholders are present.

During supervised interpretation learning, we construct the LLM input by replacing the modality placeholders with the corresponding token blocks. Let $m_{\tau}(s,x)$ denote the resulting multimodal token sequence under a sampled modality configuration $\tau$, where $\tau$ may retain the signal branch, the image branch, or both branches. We jointly optimize the modality projectors and the LLM parameters $\theta$ by minimizing the negative log-likelihood of the target response:
\begin{align}
\mathcal{L}_{c}(s, x, y_{1:T})
&= - \sum_{t=1}^{T} \log p_{\theta} \Big(
 y_t \,\Big|\, m_{\tau}(s,x), y_{1:t-1} \Big).
\end{align}

This stage provides the basis for educational explanation: when a learner submits an ECG, the model is expected to produce a response that links observable ECG evidence with diagnostic reasoning rather than only returning a label.

\begin{table*}[t]
\small
\caption{Grounded ECG Interpretation Results Independently Evaluated by GLM5.}
\centering
\resizebox{\linewidth}{!}{%
\begin{tabular}{lccccccc}
\toprule
\textbf{Metric} & \footnotesize\textbf{\makecell[l]{Diagnosis \\ Accuracy}} & \footnotesize\textbf{\makecell[c]{Analysis \\ Completeness}} & \footnotesize\textbf{\makecell[c]{Analysis \\ Relevance}} &\footnotesize\textbf{\makecell[c]{Lead \\ Evidence \\ Validity}} &\footnotesize\textbf{\makecell[c]{ECG\\Feature\\Grounding}} &\footnotesize\textbf{\makecell[c]{Evidence\\Based \\ Reasoning}}  & \footnotesize\textbf{\makecell[c]{Clinical \\ Diagnostic \\ Fidelity}} \\
\midrule
\rowcolor{gray!15}
\multicolumn{8}{c}{\footnotesize\textit{Proprietary MLLMs}} \\
\midrule
Gemini-3-Pro~\cite{google_gemini3_2025} & 14.47 & 1.85 & 0.92 & 0.72 & 38.84 & 22.18 & 39.07 \\
GPT-5.1-Instant~\cite{openai_introducing_gpt5_1_2025} & 31.75 & 2.17 & 1.75 & 1.64 & 54.54 & 37.79 & 54.19 \\
\midrule
\rowcolor{gray!15}
\multicolumn{8}{c}{\footnotesize\textit{Open-source MLLMs}} \\
\midrule
MiMo-VL-7B-SFT~\cite{XiaomiMiMo-VL-Miloco_Li_2025} & 13.98 & 1.16 & 0.55 & 0.34 & 41.84 & 17.98 & 44.51 \\
GLM-4.1V-9B-Base~\cite{GLM-4.5VGLM-4.1V-Thinking_Team_2026} & 15.27 & 0.91 & 0.48 & 0.40 & 31.19 & 16.89 & 28.68 \\
Qwen3-VL-8B-Instruct~\cite{Qwen3-VLTechnical_Bai_2025} & 20.24 & 1.62 & 0.77 & 0.36 & 39.14 & 23.02 & 38.19 \\
InternVL3-8B-Instruct~\cite{InternVL3Exploring_Zhu_2025} & 25.86 & 1.25 & 0.90 & 0.34 & 31.28 & 19.39 & 28.29 \\
MiniCPM-V-4.5~\cite{MiniCPM-V4.5_Yu_2025} & 29.24 & 1.90 & 1.36 & 0.70 & 43.23 & 27.84 & 44.94 \\
\midrule
\rowcolor{gray!15}
\multicolumn{8}{c}{\footnotesize\textit{Medical MLLMs}} \\
\midrule
MedVLM-R1~\cite{MedVLM-R1Incentivizing_Pan_2025} & 27.59 & 0.67 & 0.25 & 0.05 & 21.55 & 12.02 & 13.15 \\
Chiron-o1-8B~\cite{sun2025chirono} & 23.97 & 2.14 & 1.28 & 0.74 & 38.64 & 21.98 & 32.01 \\
QoQ-Med-VL-7B~\cite{QoQ-MedBuilding_Dai_2025} & 33.19 & 2.19 & 1.79 & 0.66 & 46.96 & 28.97 & 46.49 \\
MedGemma-4B~\cite{sellergren2025medgemma} & 35.66 & 1.23 & 0.81 & 0.09 & 36.82 & 24.41 & 33.84 \\
MedGemma-27B~\cite{sellergren2025medgemma} & 29.23 & 1.92 & 1.38 & 1.02 & 48.07 & 30.48 & 49.22 \\
HuatuoGPT-Vision-7B~\cite{HuatuoGPT-VisionInjecting_Chen_2024} & 33.81 & 2.82 & 1.89 & 0.31 & 44.85 & 31.52 & 38.21 \\
\midrule
\rowcolor{gray!15}
\multicolumn{8}{c}{\footnotesize\textit{ECG-specialized MLLMs}} \\
\midrule
PULSE~\cite{TeachMultimodal_Liu_2024} & 70.66 & 1.93 & 2.78 & 0.67 & 49.44 & 45.91 & 47.15 \\
GEM~\cite{GEMEmpowering_Lan_2025} & 78.99 & 4.05 & 4.60 & \textbf{4.87} & 74.23 & 69.17 & 71.25 \\
ECG-R1 (SFT)~\cite{ECG-R1Protocol-Guided_Jin_2026} & 83.07 & 3.99 & 4.74 & 4.53 & 86.74 & 81.06 & 88.58 \\
ECG-R1 (RL)~\cite{ECG-R1Protocol-Guided_Jin_2026} & 83.29 & \textbf{4.18} & \textbf{4.87} & 4.66 & 86.96 & 81.02 & 88.90 \\
\midrule
\textbf{UniECG} & \textbf{83.30} & 4.08 & 4.83 & 4.62 & \textbf{86.98} & \textbf{81.42} & \textbf{88.91} \\
\bottomrule
\end{tabular}%
}
\label{tab:grounded_interpretation}
\end{table*}

\subsection{Learning Controllable ECG Signal Generation}

The second stage extends the language model with ECG generation ability. Inspired by GILL~\cite{GeneratingImages_Koh_2023}, we introduce an ECG generation start token, \ecgstart, and a set of ECG latent tokens, $\ecgtok{0}, \ecgtok{1}, \ldots, \ecgtok{n-1}$, to equip the language model with the ability to generate ECG signals. Here, $n$ denotes the number of ECG latent tokens and is fixed at $8$ in our experiments. During training, we exclusively update the token embedding matrix, the language modeling head, the final normalization layer, the last eight Transformer blocks, and a lightweight ECG-to-DiffuSETS mapper, while keeping the remaining backbone parameters frozen.

\begin{table*}[t]
\centering
\caption{ECG Generation Results.}
\label{tab:generation_results}
\small
\resizebox{0.88\textwidth}{!}{%
\begin{tabular}{l|cccc|c|c|c}
\toprule
 & \multicolumn{4}{c|}{\textbf{Signal Level}} & \textbf{Feature Level} & \textbf{Diagnostic Level} & \textbf{Success} \\
\textbf{Metric} & \textbf{FID} & \textbf{Precision} & \textbf{Recall} & \textbf{F1} & \textbf{HR MAE } & \textbf{CLIP} & \textbf{Rate} \\
\midrule
DiffuSETS~\cite{DiffuSETS12-Lead_Lai_2025} & \textbf{25.26} & \textbf{0.941} & \textbf{0.709} & \textbf{0.809} & \textbf{8.70} & \textbf{0.746} & \textbf{100\%} \\
UniECG & 79.43 & 0.939 & 0.615 & 0.743 & 11.72 & 0.732 & 95.61\% \\
\bottomrule
\end{tabular}%
}
\end{table*}

The model is trained to generate the target ECG-token sequence when the input requests ECG synthesis. This objective supervises the transition from natural-language response generation to ECG-token generation by requiring the model to emit the ECG-start token before producing the ECG tokens:
\begin{align}
\mathcal{L}_{e}(y)
&= - \lambda_{\mathrm{s}}
\log p_{\theta, E_{\mathrm{ECG}}}
\big([\mathrm{ECG\_START}] \mid y\big)
\nonumber\\
&\quad
- \lambda_{\mathrm{f}}
\log p_{\theta, E_{\mathrm{ECG}}}
\big([\mathrm{ECG}_0] \mid y,[\mathrm{ECG\_START}]\big)
\nonumber\\
&\quad
- \lambda_{\mathrm{r}}
\sum_{i=1}^{n-1}
\log p_{\theta, E_{\mathrm{ECG}}}
\big(
[\mathrm{ECG}_i]
\mid y,[\mathrm{ECG\_START}],[\mathrm{ECG}_{<i}]
\big).
\end{align}

For training data construction, textual reports from MIMIC-IV-ECG~\cite{MIMIC-IV-ECGDiagnostic_Gow_2023} are converted into paired query--answer examples. A query takes the form ``Generating an ECG of ...'', and the assistant's response consists of a sequence of ECG tokens initiated by \ecgstart. This procedure relies on text reports for learning the generation trigger and semantic condition.

To synthesize ECG signals, UniECG adopts DiffuSETS~\cite{DiffuSETS12-Lead_Lai_2025}, a diffusion-based framework for text-conditioned 12-lead ECG generation. The key step is to map the hidden states of $\ecgtok{0}, \ldots, \ecgtok{n-1}$ into the DiffuSETS text embedding space. We implement this mapping with a lightweight encoder--decoder Transformer $f_{\omega}$, which takes the ECG-token hidden states and $L$ learnable query embeddings $q_{1:L}$ as input. The mapper is trained to project the ECG token representations into the DiffuSETS text embedding space. To this end, we minimize the mean squared error between the mapper output and the frozen DiffuSETS text embedding $T_{\psi}(\tilde y)$. Here, $\tilde y$ denotes the report text reformatted according to the DiffuSETS input format and is used to construct the teacher embedding:
\begin{align}
\mathcal{L}_{d}(y)
&= \Big\| f_{\omega}\big(
    h_{\{\theta \cup E_{\mathrm{ECG}}\}}(y,[\mathrm{ECG}_0]), \ldots, \nonumber\\
&\quad h_{\{\theta \cup E_{\mathrm{ECG}}\}}(y,[\mathrm{ECG}_{n-1}]), q_{1:L} \big) 
- T_{\psi}(\tilde y) \Big\|_{2}^{2}.
\end{align}

The overall objective of the second stage is
\begin{align}
\mathcal{L}_{g}(y) = \mathcal{L}_{e}(y) + \mathcal{L}_{d}(y).
\end{align}

\subsection{Interactive Educational Inference}

At inference time, UniECG selects its response mode according to the user's request. For ECG explanation, the model receives ECG data and produces an evidence-based textual response. For ECG generation, the model outputs special ECG tokens, which are passed through the mapper and then decoded by the pretrained DiffuSETS decoder to synthesize a signal-level ECG. This design supports an educational workflow in which a learner can alternate between asking why an ECG indicates a condition and asking what a requested condition may look like.

\section{Experiments}

We evaluate UniECG along the two educational capabilities targeted by the framework: evidence-based ECG explanation and controllable ECG case generation. For explanation, we compare UniECG with proprietary, open-source, and medical-domain multimodal language models on grounded ECG interpretation metrics. For generation, we compare the UniECG token-conditioned generation pathway with the direct text-conditioned DiffuSETS pathway.

\subsection{Experimental Setup}

For the interpretation stage, UniECG follows the training and evaluation protocol of ECG-R1~\cite{ECG-R1Protocol-Guided_Jin_2026}, and directly adopts the ECG-R1-8B-RL weights as the model initialization. For the generation stage, we construct data from MIMIC-IV-ECG~\cite{MIMIC-IV-ECGDiagnostic_Gow_2023} after excluding the first 50,000 records and the 2,381 records used in the GEM test set, leaving 737,875 training samples. Each generation query is formatted as ``Generating an ECG of [Reports]'', where multiple diagnostic statements are concatenated with commas. The supervision follows the original DiffuSETS text preprocessing: diagnostic reports are converted into structured descriptions and encoded into 768-dimensional text embeddings for cross-modal alignment.

During Stage II training, UniECG generates \ecgstart{} followed by $\ecgtok{0}$--$\ecgtok{7}$ and uses a lightweight generation mapper to project the corresponding hidden states into the DiffuSETS text-conditioning space. We update only the token embeddings, the language-modeling head, the generation mapper, the final LayerNorm layer, and the last eight Transformer blocks, while keeping all remaining parameters frozen. Training is performed for three epochs on three NVIDIA A100 GPUs with per-device batch size 1, gradient accumulation of 2 steps, learning rate $2\times10^{-5}$, weight decay 0.01, maximum sequence length 1024, and gradient checkpointing enabled.

For evaluation, grounded ECG explanation is tested on the 2,381-sample GEM test set and scored by GLM-5 under the ECG-R1 rubric. ECG signal generation is evaluated on 10,000 records randomly sampled from the held-out first 50,000 MIMIC-IV-ECG records. Both DiffuSETS and UniECG are evaluated under the original DiffuSETS protocol to ensure comparability.

\subsection{Grounded ECG Explanation}

Table~\ref{tab:grounded_interpretation} reports grounded ECG interpretation results on 2,381 ECG cases using an LLM-based evaluator. UniECG is initialized from ECG-R1 and further trained to support the closed-loop educational workflow. The results show that this additional training does not substantially damage the original ECG interpretation ability: UniECG remains close to ECG-R1 on analysis completeness, analysis relevance, and lead evidence validity, and achieves the best scores on diagnosis accuracy, ECG feature grounding, evidence-based reasoning, and clinical diagnostic fidelity.

Compared with proprietary, open-source, and general medical-domain MLLMs, UniECG also retains a clear advantage. This suggests that the model can add the generation-oriented educational interface while preserving the specialized ECG reasoning learned from prior ECG-focused training. For the proposed education setting, this preservation is important because the system must generate teaching cases without sacrificing its ability to explain ECG evidence.

\subsection{Controllable ECG Case Generation}

Table~\ref{tab:generation_results} compares UniECG with direct DiffuSETS conditioning on signal-, feature-, and diagnostic-level generation metrics. Direct DiffuSETS is an upper reference for this experiment because it receives the input text in the native conditioning format of the pretrained generator. In contrast, UniECG must first decide to enter the ECG generation mode, produce special ECG tokens, and map their hidden states into the DiffuSETS text-conditioning space. Under this more difficult pathway, UniECG achieves a 95.61\% decoding success rate, showing that most textual learning objectives can be converted into a valid signal-generation process through the learned ECG-token interface.

The quality metrics show that the generated cases are usable as preliminary synthetic teaching examples, but still lag behind direct generator conditioning. UniECG obtains signal-level precision close to DiffuSETS (0.939 versus 0.941), suggesting that many generated signals remain within plausible ECG regions. However, its FID increases from 25.26 to 79.43, recall decreases from 0.709 to 0.615, and F1 decreases from 0.809 to 0.743, indicating reduced distributional coverage and weaker signal-level fidelity. At the feature and diagnostic levels, the HR MAE increases from 8.70 to 11.72 and CLIP decreases from 0.746 to 0.732. These gaps suggest that the main bottleneck is not only waveform decoding, but also semantic preservation when language-model hidden states are compressed into the generator-conditioning space. Improving this alignment is therefore central to making closed-loop ECG teaching-case generation more reliable.

\section{Future Work}

Future work will evaluate UniECG with medical students in realistic ECG education settings. We plan to collect human feedback on whether the closed-loop workflow helps students understand ECG evidence, review generated cases, and practice targeted ECG reasoning.

\section{Conclusion}

This paper presents UniECG as a step toward a closed-loop educational system for ECG learning. UniECG connects evidence-based explanation with controllable case generation, allowing learners to move between interpreting observed ECGs and constructing synthetic examples from textual learning goals. Experiments show strong grounded explanation performance and demonstrate the feasibility of token-based ECG case generation, while also indicating the need for further evaluation with medical students.

\bibliographystyle{ACM-Reference-Format}
\bibliography{uniecg-new}

\appendix

\twocolumn[{%
\section{Qualitative Examples of UniECG's Dual Capabilities}
\label{app:case_studies}

\begin{center}
  \includegraphics[
    width=\textwidth,
    height=\textheight,
    keepaspectratio
  ]{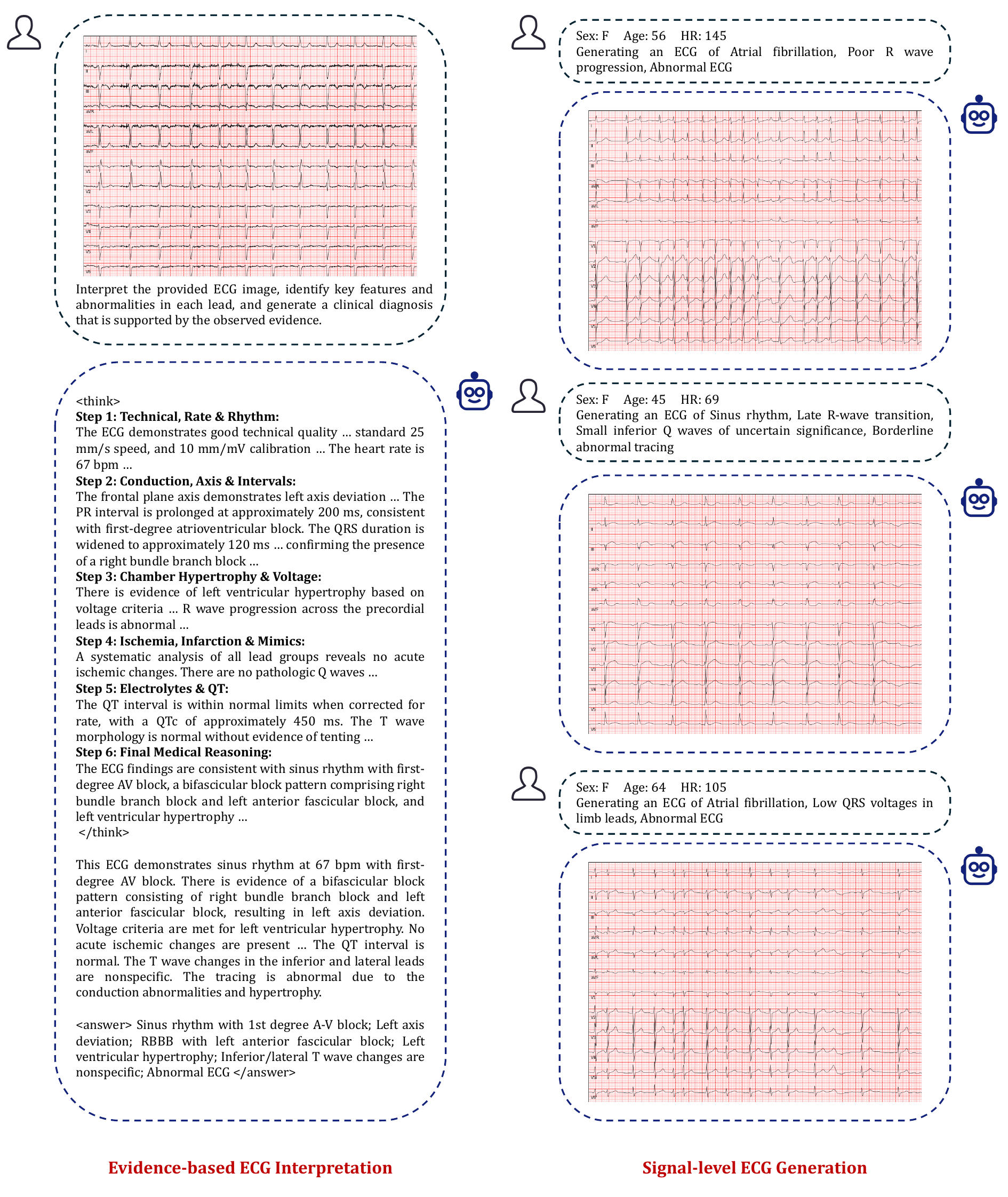}

  \captionof{figure}{Qualitative examples of UniECG's dual capabilities. The left panel shows evidence-based ECG interpretation from an observed ECG image, while the right panel shows text-conditioned ECG generation under different diagnostic conditions.}
  \label{fig:appendix_cases}
\end{center}
\vspace{1em}
}]

\end{document}